# A SAM-guided and Match-based Semi-Supervised Segmentation Framework for Medical Imaging

**Running Title:**   SAMatch for Semi-supervised Segmentation

Guoping Xu[1]
Xiaoxue Qian[1]
Hua-Chieh Shao[1]
Jax Luo[2]
Weiguo Lu[1]
You Zhang[1]

[1]The Medical Artificial Intelligence and Automation (MAIA) Laboratory
Department of Radiation Oncology
University of Texas Southwestern Medical Center, Dallas, TX 75390, USA
[2] Massachusetts General Hospital, Harvard Medical School, Boston, MA 02115, USA

Corresponding address:

You Zhang
Department of Radiation Oncology
University of Texas Southwestern Medical Center
2280 Inwood Road
Dallas, TX 75390
Email: You.Zhang@UTSouthwestern.edu
Tel: (214) 645-2699

# Abstract

**Background:** Semi-supervised segmentation leverages sparse annotation information to learn rich representations from combined labeled and label-less data for segmentation tasks. The Match-based framework, by using the consistency constraint of segmentation results from different models/augmented label-less inputs, is found effective in semi-supervised learning. This approach, however, is challenged by the low quality of pseudo-labels generated as intermediate products for training the network, due to the lack of the 'ground-truth' reference.

**Purpose:** This study aims to leverage the foundation model, segment anything model (SAM), to assist unsupervised learning of Match-based frameworks. Trained with an extremely large dataset, SAM-based methods generalize better than traditional models to various imaging domains/tasks, allow it to serve as an assistant to Match-based frameworks to improve the quality of intermediate pseudo-labels for un-supervised learning.

**Methods:** We propose SAMatch, a SAM-guided Match-based framework for semi-supervised medical image segmentation. Our approach involves two main steps: First, we use pre-trained Match-based models to extract high-confidence predictions for prompt generation. Second, these prompts and unlabeled images are input into a fine-tuned SAM-based method to produce high-quality masks as pseudo-labels. And the refined pseudo-labels are further fed back to train the Match-based framework. SAMatch can be trained in an end-to-end manner, facilitating interactions between the SAM- and Match-based models.

**Results:** SAMatch demonstrates robust performance across multiple datasets. We achieved state-of-the-art results on the ACDC cardiac MRI dataset, the BUSI breast ultrasound dataset, and an in-house MRLiver dataset. On ACDC, with only three labeled cases, we achieved a Dice score of 89.36% on 20 testing cases. For BUSI, with just 30 labeled samples, we attained a Dice score of 77.76% on 170 testing samples. On MRLiver, training with only three labeled cases resulted in a Dice score of 80.04% on 12 testing scans.

**Conclusions:** Our SAMatch framework shows promising results in semi-supervised semantic segmentation, effectively tackling the challenges of automatic prompt generation for SAM and high-quality pseudo-label generation for Match-based models. It can help accelerate the adoption of semi-supervised learning in segmentation tasks, particularly in data-scarce scenarios. Our data and code will be made available at https://github.com/apple1986/SAMatch.

# 1 Introduction

Semantic segmentation is crucial in medical image analysis, aiding disease diagnosis and treatment planning by outlining key anatomical structures and abnormalities [1]. Manual segmentation is a laborious process that requires substantial human effort. Automatic segmentation reduces the need for manual contouring and mitigates inter-operator discrepancies while enhancing the efficiency of clinical workflows [2]. Recently, automatic segmentation has seen tremendous progress since the development of deep neural networks, such as FCN [3], U-Net [4], and DeepLab [5]. These neural networks are usually developed in a supervised fashion, by using the 'gold-standard' human segmentations as reference labels to train the network to generalize to unseen test samples. However, these fully supervised methods usually require many annotated labels for training, which is time-consuming and expensive to collate. In contrast, semi-supervised segmentation, which aims to leverage a few labeled images to train a model along with many unlabeled images, has a great potential for annotation-efficient learning [6] in data-scarce clinical scenarios.

For semi-supervised learning, the Match-based framework via output consistency regularization has been widely adopted due to its simplicity and effectiveness [7-9]. Such a framework trains a network to yield consistent predictions when perturbating the input/model, provided that the true label remains unchanged [10]. One of the representative works is FixMatch [11]. FixMatch first generates pseudo labels from weakly-augmented inputs and retains those labels with a high-confidence prediction. These retained pseudo labels are used as target reference labels for strongly-augmented versions of the same inputs. The reference labels and the strongly-augmented inputs are used to further train the model by enforcing a consistency regularization. Adopting the principle of FixMatch, many Match-based methods are explored for semi-supervised segmentation using weakly-augmented and strongly-augmented samples, such as mixup in CCAT-NET [12], copy-paste in BCP [13], and elastic deformations used in a Siamese architecture [14], etc. Recently, UniMatch was also proposed to augment the inputs from both the image level and the feature level, to leverage a broader perturbation space to enforce consistency regularization and promote label-less learning effectively [9].

Most of these Match-based methods generally rely on high-quality pseudo labels to enact the consistency loss between weakly-augmented and strongly-augmented images, by viewing the pseudo labels as the 'ground truth' to train the models [15]. However, there is no guarantee that the model will generate high-quality pseudo labels from weakly-augmented images, and poor-quality labels will break the consistency assumption to reduce the model's accuracy. For example, small perturbations in weakly-augmented images may introduce noise or outliers, causing the model to focus on these noisy regions and produce low-quality pseudo labels. The choice of augmentation strategies for both weakly- and strongly-augmented images also requires careful consideration to maintain label consistency in training [6, 8]. Resultingly, the key question for Match-based methods is: *How to generate high-quality pseudo labels for unlabeled images during the training stage, for semi-supervised segmentation under the Match-based framework?*

Recently, the 'segment anything model' (SAM) was introduced as a foundation model for generic object segmentation [16]. It was trained on 11 million 2D natural images and over 1 billion segmentation masks. Experimental results demonstrate its strong segmentation performance with proper prompts such as points, boxes, masks, or texts. Due to its powerful generalization ability, many SAM-based works have been proposed for medical image segmentation [17, 18]. In [19], a tailored SAM model (LeSAM) was proposed for lesion segmentation, which integrates the general knowledge from pre-trained SAM with medical domain knowledge via a lightweight U-shaped network. MedSAM was proposed in [20], which was fine-tuned on a large-scale medical image dataset and demonstrated better accuracy and robustness than modality-specific traditional models. However, these SAM-based models require input prompts (points, boxes, etc.) to denote the location of segmentation targets, and the manual nature of the prompts hinders their seamless application in real-world clinical scenarios. In addition, we found that these SAM-based models may still require further fine-tuning for specific tasks. For example, Table 1 shows the performance of SAM and MedSAM before and after fine-tuning on an in-house liver segmentation dataset, which demonstrates that the accuracy of both models was boosted after fine-tuning.

**Table 1.** Comparison of segmentation results for SAM-based models (before and after fine-tuning) on an in-house liver dataset. The prompts are based on the 'ground-truth' masks and we used 3 labeled cases for fine-tuning.

| Method | Prompt | Mean Dice (%) |
|---|---|---|
| SAM | Point | 28.47 |
| SAM-finetune | | 77.76 |
| MedSAM | Box | 78.96 |
| MedSAM-finetune | | 90.61 |

The respective advantages and limitations of Match-based and SAM-based methods render them complementary with potential gains if combined (see Table 2). While Match-based models may struggle to generate high-quality pseudo labels exactly matching the whole structures, they can automatically produce relatively high-confidence prompts—such as points or boxes—after a few epochs of training, which is a boon for SAM. Conversely, with prompts from Match-based methods, SAM-based models can generate high-quality prediction masks after few-sample fine-tuning, which can in turn serve as high-quality pseudo labels to effectively train Match-based methods.

**Table 2.** Comparison of the respective advantages and limitations of Match-based and SAM-based models.

| Method | Advantage | Limitation |
| --- | --- | --- |
| Match-based | Automatically generate relatively high-confidence prompts from prediction results | Require high-quality pseudo labels for effective training |
| SAM-based | Produce high-quality prediction masks when given appropriate prompts | Rely on manually-provided, task-related prompts; Need fine-tuning |

To leverage the strengths of both Match-based and SAM-based methods while addressing their respective limitations, we propose a novel framework named SAMatch in this study. In SAMatch, Match-based models generate prompts to serve as inputs for SAM-based models. Based on the prompts, SAM-based methods generate predicted masks and feed them back into Match-based models as pseudo labels. The Match-based models are trained in synergy with the fine-tuning of the SAM-based methods, to maximize the accuracy of semi-supervised learning. The key distinction between the traditional Match-based methods and the SAMatch framework is illustrated in Figure 1. In SAMatch, the fine-tuned SAM-based methods serve to assist the teacher model of Match-based methods.

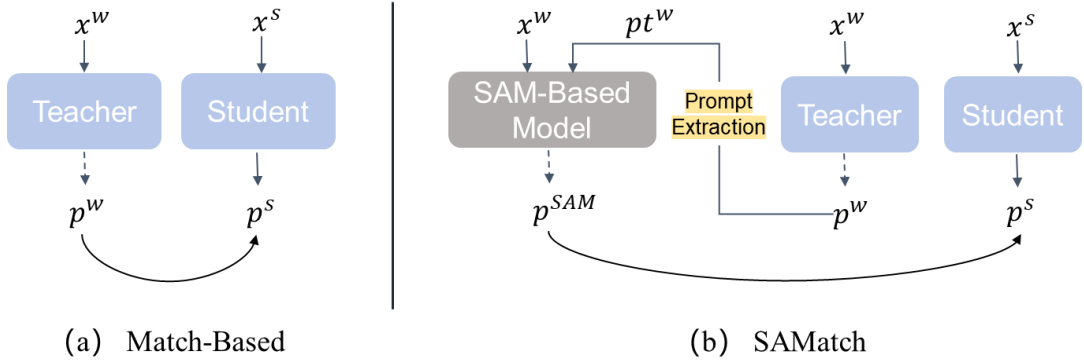

(a) Match-Based      (b) SAMatch

**Figure 1**. Comparison between (a) the Match-based methods and (b) the proposed SAMatch framework. Here, the $x^w$ and $x^s$ denote the weakly-augmented and strongly-augmented inputs. The $p^w$ and $p^s$ represent the corresponding prediction outputs. $pt^w$ denotes the extracted prompts from $p^w$, and $p^{SAM}$ denotes the predicted pseudo labels from the SAM-based model. The solid curves represent supervision from the pseudo labels for training Match-based models. The dashed arrows linking the teacher model and $p^w$ indicate that the teacher model is not updated through network backpropagation. Instead, their network weights are updated as exponential moving averages of the student networks' weights.

In summary, our main contributions to this study are as follows:

- We introduce a unified semi-supervised framework (SAMatch) that integrates SAM-based methods with Match-based semi-supervised methods, leveraging their advantages to address their respective limitations. SAMatch is a general framework that can apply to various versions of SAM-based and Match-based methods, which can be tailored to the needs of specific segmentation tasks.
- We develop a strategy for the automatic generation of prompts using Match-based semi-supervised methods, thereby eliminating the need for manual prompting to drive SAM-based subcomponents. We trained Match-based methods and fine-tuned SAM-based methods within the SAMatch framework, allowing them to adapt to the specific segmentation task and achieve better performance.
- We demonstrate through comprehensive evaluations that our framework outperforms existing semi-supervised methods by a significant margin, particularly in few-shot learning scenarios with extremely limited labels.

## 2 Related Work

### 2.1 SAM-based segmentation methods

SAM has demonstrated impressive results as a foundational model for general image segmentation [16]. Due to the expensive and time-consuming nature of labeling medical images, SAM has garnered significant attention in medical image segmentation. In [17] and [18], SAM was applied to medical images, showing its remarkable performance with appropriate prompts. However, SAM's performance can be unstable due to challenges such as low image contrast, boundary fuzziness, and structural complexity. In [20], a large-scale medical image dataset comprising over one million image-mask pairs was created and used to fine-tune the original SAM, resulting in MedSAM. This adaptation led to enhanced accuracy and robustness compared to earlier task-specific segmentation models, highlighting the importance of fine-tuning SAM for better performance in medical image segmentation.

While fine-tuned SAM variations showed better results, existing adaptations require appropriate prompts (such as points and boxes) as inputs to achieve satisfactory segmentation performance [7]. In many scenarios, these prompts are directly derived from 'ground-truth' labels during testing [20, 21], which is often impractical in clinical settings. In [22], the YOLOv8 object detection model was trained to provide bounding boxes as prompts for SAM. AutoSAM [23] introduces a new encoder to generate surrogate prompts for SAM, demonstrating improved performance across various benchmarks. In AutoSAMUS [24], an auto prompt generator was proposed to automatically create prompt embeddings for ultrasound image segmentation. While these methods effectively generate prompts to automate SAM, a significant limitation is their reliance on a substantial amount of labeled data for supervised learning. Some studies [25-27] explored applying SAM to semi-supervised or weakly-supervised learning scenarios. However, these approaches typically utilize SAM solely as a generator of pseudo labels, which can lead to poor label quality without fine-tuning. In this study, we propose a framework that uses fine-tuned SAM-based models to generate pseudo labels to train Match-based semi-supervised models. The fine-tuning of the SAM-based models and the training of Match-based methods are designed in synergy to boost the accuracy of both components to maximize the overall segmentation quality. This approach not only improves the quality of pseudo labels generated by SAM-based models but enhances the segmentation performance of the Match-based model within a semi-supervised learning context.

**2.2 Match-based semi-supervised learning**

Match-based methods are grounded on the consistency hypothesis, which posits that controlled perturbations to data or model should not alter the model's predictions. The basic Match-based framework feeds the same input into different models to enforce the consistency of their outputs. Notable examples include the $\Pi$ model [28] and the Mean Teacher framework (Figure 2) [29, 30]. The $\Pi$ model employs a dropout strategy on the same model to generate two predictions, and their differences are used as an unsupervised loss assuming consistent outputs for the same input image. The Mean Teacher framework consists of two models (teacher and student) with identical architectures and uses exponential moving averages of the student model's weights to update the teacher model. The teacher model's predictions serve as pseudo labels to train the student model. Additionally, approaches that incorporate structural consistency [15], interpolation consistency [31], uncertainty-aware consistency [30], uncertainty-rectified pyramid consistency [32], and unreliable pseudo-label consistency [33] were developed to improve the alignment between teacher and student models. However, a significant drawback of these methods is the tight coupling between student and teacher models, as they often share the same architecture [34]. This can hinder the generation of high-quality pseudo labels if one of the models fails to train effectively.

To better decouple the teacher and the student networks for more efficient training, data-level differential augmentations were further introduced to augment the input images with different augmentations before feeding them into the teacher and the student models. Such a strategy typically employs weak-to-strong data augmentations, a concept popularized by FixMatch [11]. FixMatch focuses on image-level perturbations of inputs and uses labels with a high-confidence prediction from the teacher model as pseudo-labels for training the student model. In contrast, UniMatch [9] unifies image-level and feature-level perturbations to exploit a broader augmentation space. Additionally, FlexMatch [35] employs a curriculum learning approach, while FreeMatch [36] introduces a self-adaptive method to explore unlabeled data more effectively than FixMatch. In general, differential data-level augmentations allow Match-based methods to capitalize on rich information of input images by employing various augmentations, such as Cutout [37], CutMix [38], adaptive CutMix [39], and bidirectional copy-paste [13], to promote consistency learning. However, a lingering limitation is that the predictions from unlabeled data, used as pseudo labels, are not always accurate or consistent with labeled data. It leads to noise or mislabeling, resulting in suboptimal performance for semi-supervised learning [8].

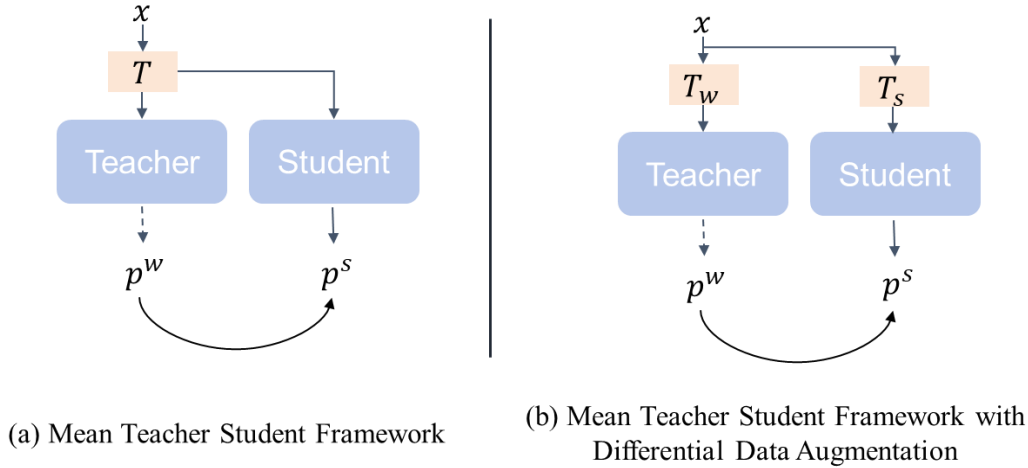

(a) Mean Teacher Student Framework

(b) Mean Teacher Student Framework with Differential Data Augmentation

**Figure 2**. (a) The basic mean teacher-student framework, which inputs the same augmented ($T$) image into different models to enforce consistency regularization. (b) The mean teacher-student framework with differential data augmentations. It uses differently-augmented images (by weak augmentation $T_w$ and strong augmentation $T_s$) to better decouple the student and teacher models for more effective training.

In summary, current Match-based methods cannot guarantee the generation of high-quality predictions as pseudo labels for training under the consistency assumption. Our proposed framework, SAMatch, integrates SAM-based models into the Match-based framework, aiming to produce high-quality labels that alleviate the limitations of training with unlabeled data. The SAM-based model functions as an assistant to the teacher model in the mean teacher-student framework, addressing the low-quality pseudo-label issue.

## 3 Materials and Methods

Semi-supervised segmentation aims to fully explore unlabeled images with a limited number of labeled images. Our method is the extension of Match-based frameworks, such as FixMatch and UniMatch. We first review the preliminaries and the basic architecture of Match-based, semi-supervised segmentation methods in Section 3.1. The details of SAMatch are introduced in Section 3.2, and we summarize the training and testing procedures of SAMatch in Sections 3.3-3.6.

**3.1 Preliminaries**

For semi-supervised segmentation tasks, there is a labeled image dataset $(x_l, y_l)$ and an unlabeled image dataset $x_u$, while the size of the unlabeled dataset is typically much larger than that of the labeled dataset. In the training stage of Match-based methods with a teacher-student framework and data-level differential augmentation (Figure 3), two different augmentation strategies are applied on the same input image to yield two augmented images: weak augmentation $A_w$ and strong augmentation $A_s$. Specifically, $A_w$ includes random angle rotation and cropping, while $A_s$ involves the same random transformation as $A_w$ with additional strong perturbations, such as color jittering or noise infusion. Subsequently, the weakly-augmented images $A_w(x_u)$ are used to create pseudo-labels $y_u$ with the teacher model. The pseudo-labels are then used in training the student model where the strongly-augmented images $A_s(x_u)$ are fed as input.

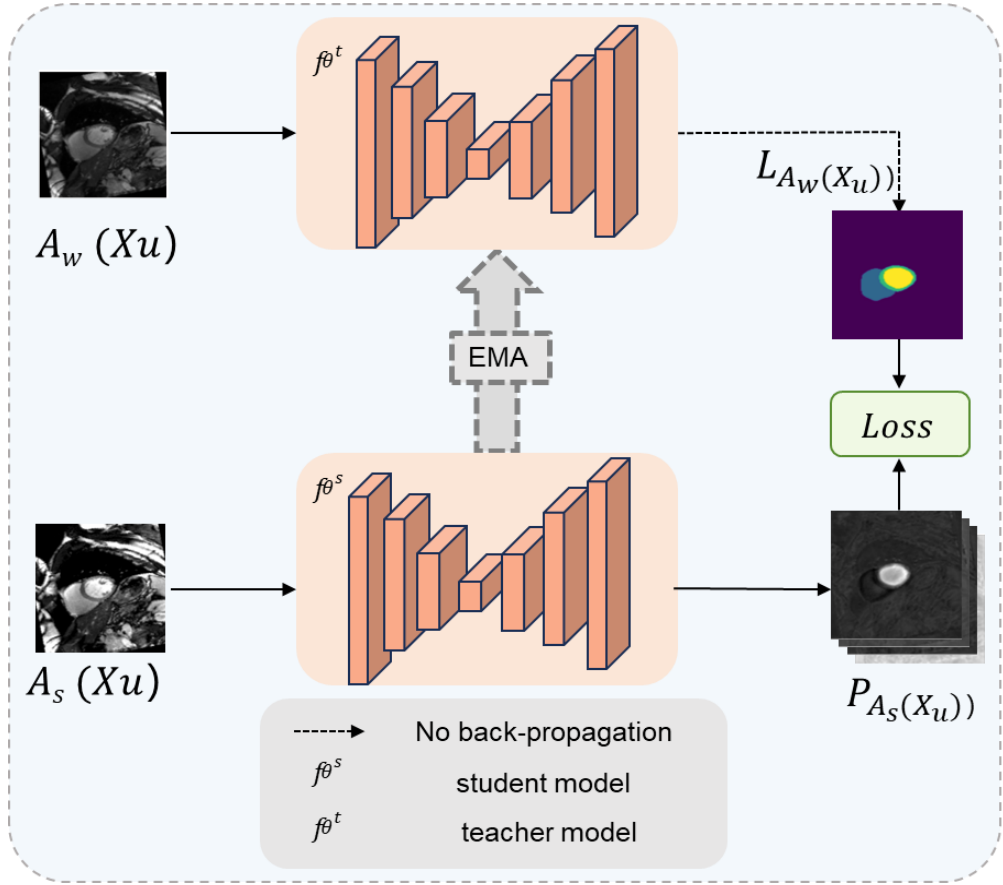

**Figure 3**. The architecture of Match-based, semi-supervised segmentation based on a mean teacher-student framework. The teacher model generates pseudo labels which are used for training the student model. The supervised training component, which was based on labeled images, was omitted from the schematic drawing for simplicity. $L_{A_w(x_u)}$ indicates the pseudo label created by the teacher network, with high-confidence prediction thresholding. $P_{A_s(x_u)}$ denotes the student network's prediction. EMA: exponential moving average.

Here, the generation of pseudo labels in Match-based methods is represented as:

$$y_u = argmax(softmax(f_{\theta^t}(A_w(x_u))) > T), \quad (1)$$

where $f_{\theta^t}$ represents the teacher network and its weights come from the student model as exponential moving averages. $T$ is a pre-defined threshold value (0.95) to select pixels with high-confidence predictions as pseudo labels. The loss of unlabeled data $\mathcal{L}_u$ is as follows:

$$\mathcal{L}_u = \mathcal{L}_{DICE}(f_{\theta^s}(A_s(x_u)), y_u) + \mathcal{L}_{ce}(f_{\theta^s}(A_s(x_u)), y_u), \quad (2)$$

where $\mathcal{L}_{DICE}$ and $\mathcal{L}_{ce}$ denote the DICE loss and the cross-entropy loss, respectively. $f_{\theta^s}$ represents the student network. In addition to the unsupervised loss, for the available labeled samples, we also compute the supervised loss $\mathcal{L}_l$ using DICE and cross-entropy:

$$\mathcal{L}_l = \mathcal{L}_{DICE}(f_{\theta^s}(A_s(x_l)), y_l) + \mathcal{L}_{ce}(f_{\theta^s}(A_s(x_l)), y_l). \quad (3)$$

For the supervised training component, the labeled images are directly input into the student network after augmentation without going through the teacher model-based pseudo-label generation process, and the output is directly used to compute the supervised loss to update the student network. The total loss function of the Match-based model can be expressed as:

$$\mathcal{L} = \mathcal{L}_l + \lambda \mathcal{L}_u, \quad (4)$$

where $\lambda$ is a weighting coefficient balancing the supervised and unsupervised losses. In this study, we take the same strategy as [28] to adaptively update $\lambda$.

**3.2 Proposed SAMatch framework**

Our proposed SAM-guided and Match-based framework consists of two primary components: the teacher and student models for the Match-based network, and the SAM-based network. The overall architecture is illustrated in Figure 4.

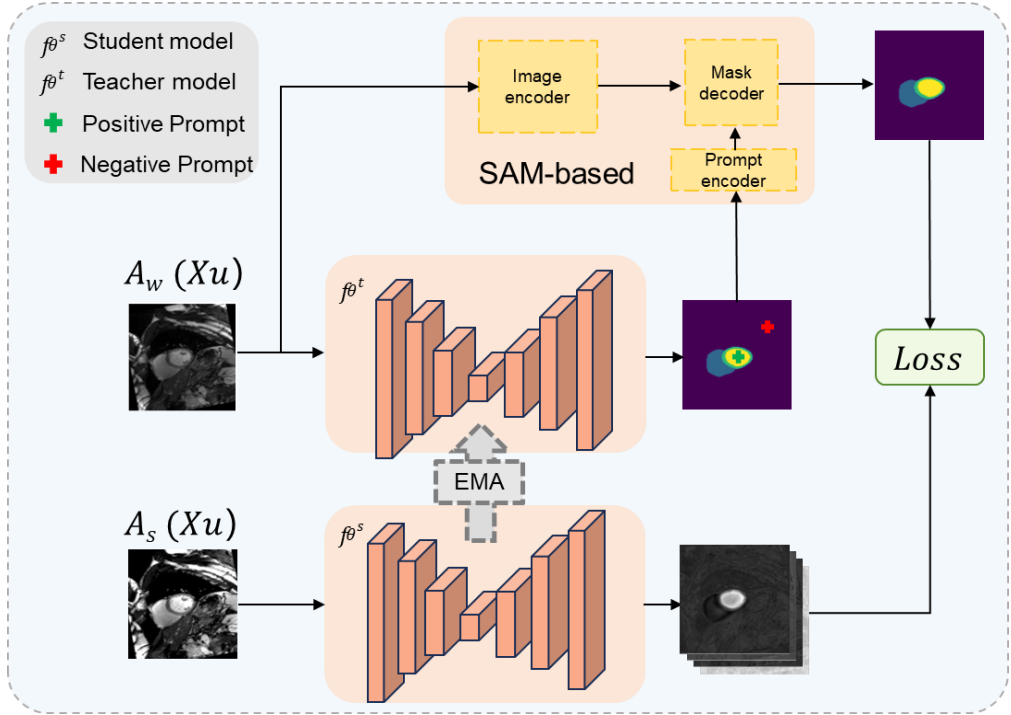

**Figure 4**. Overall SAMatch framework. The framework consists of two main components: the teacher and student models, and the SAM-based network. A teacher model generates prediction masks from weakly-augmented images, from which prompts (e.g., points and boxes) are extracted. The SAM-based network then creates high-quality pseudo labels. The student model is trained with the SAM-based pseudo labels and the strongly-augmented images. The supervised training component, which was based on labeled images, was omitted from the schematic drawing for simplicity. EMA: exponential moving average.

Specifically, SAMatch adopts a similar teacher-student model architecture (Fig. 3) and incorporates different image augmentation strategies to generate pseudo labels and improve the effectiveness of unsupervised learning. First, the input image is weakly augmented ($A_w(x_u)$) and then fed into the teacher model $f_{\theta^t}$ to generate pseudo-labels. The pseudo-labels were then processed to generate prompts to feed into the SAM-based network, along with the weakly-augmented image, to enhance the quality of the pseudo-labels. The enhanced pseudo-labels are then used to supervise the training of the student model $f_{\theta^s}$, using the strongly-augmented image ($A_s(x_u)$) as input. An unsupervised loss function $\mathcal{L}_u$, as defined in Eq. 2, is employed to measure the difference between the student model's output and the pseudo-labels. Throughout the training process, the teacher model adopts a no-backpropagation strategy, where its primary role is to generate coarse pseudo-labels rather than directly participating in learning, and its weightings were updated from those of the student model using the exponential moving average.

### 3.3 Training of SAMatch

SAMatch involves two main training steps, as provided in algorithm 1.

**Algorithm 1: Training of SAMatch.**

Hyperparameter Settings: learning rates, number of warm-up steps $M$ (for updating the Match-based network and the SAM-based network separately), and number of interactive steps $N$.

*for M (warm-up) steps do*

1) Random augmentation strategy for each batch:
- Augment labeled data: $A_s(x_l)$ and get prompts from the corresponding labels
- Transform unlabeled data with weak and strong augmentations: $A_w(x_u)$ and $A_s(x_u)$
2) Update the Match-based student network with labeled data $A_s(x_l)$ as well as unlabeled data $A_s(x_u)$ via pseudo-labels (generated from the teacher network with input $A_w(x_u)$); and update the teacher network by the exponential moving average of the student network's weightings
3) Fine-tune the SAM-based network with labeled data $A_s(x_l)$ and prompts

*end for*

*for N (interactive) steps do*

1) Random augmentation strategy for each batch:
- Augment labeled data: $A_s(x_l)$ and get prompts from the corresponding labels
- Transform unlabeled data with weak and strong augmentations: $A_w(x_u)$ and $A_s(x_u)$
2) Input weakly-augmented unlabeled data $A_w(x_u)$ to the teacher model, and get prompts from the prediction mask $p^w$
3) Input $A_w(x_u)$ and prompts to SAM-based network, and get the pseudo-labels $p^{SAM}$
4) Update the student model with strongly-augmented data $A_s(x_u)$ and pseudo-labels $p^{SAM}$
5) Update the teacher network by the exponential moving average of the student network's weightings

*end for*

The student model training of the Match-based network involves a small set of labeled data and a large set of unlabeled data. For labeled data, the model is trained via supervised loss with 'ground-truth' labels; For unlabeled data, the model leverages the pseudo-labels generated by the teacher model (during the warm-up steps) or by the SAM-based network (during the interactive steps). By combining direct supervision from labeled data with the guidance of pseudo-labels from unlabeled data, the student model can better utilize the limited annotated data and improve its generalizability to diverse scenarios. On the other hand, SAM is a semantic segmentation network with strong generalization capabilities. However, due to the significant differences in characteristics between natural and medical images, directly applying SAM to medical image segmentation may yield suboptimal results. Therefore, it is necessary to fine-tune SAM to adapt it for medical image segmentation tasks. In SAMatch, the SAM-based network is fine-tuned in the warm-up steps through labeled data (Algorithm 1).

### 3.4 Evaluation datasets

In this study, we used three datasets for evaluation:

**ACDC:** The ACDC dataset comprises 150 cardiac magnetic resonance imaging cases. In each case, we used two scans of different cardiac motion states: end-diastolic and end-systolic. Among these, 100 cases have publicly-available manual annotations and were used for our training and evaluation. Same as the partitioning strategy outlined in LeViT-UNet [40], the annotated cases were divided into 70 for training, 10 for validation, and 20 for testing. For the semi-supervised segmentation training, the 70 training cases were further divided: we conducted experiments using either 1 labeled case or 3 labeled cases for supervised training, with the remaining 69 and 67 cases employed for unsupervised training (with labels removed). As for the scenario of 1 labeled case, we only used the end-diastolic scan for supervised learning, to evaluate the models' performance with extremely limited labels for training. The segmentation task aims to segment three structures: the left ventricle (LV), right ventricle (RV), and myocardium (MYO).

**BUSI:** The BUSI dataset comprises 780 breast ultrasound images collected from 600 female patients aged 25 to 75. We excluded 233 normal cases without lesions, leaving 547 samples that include both benign and malignant lesions for segmentation. The 547 samples were partitioned into 330, 47, and 170 cases for training, validation, and testing, following the splitting methodology used in [41]. In the semi-supervised segmentation task, 10 labeled and 320 unlabeled cases were further partitioned from the training dataset. Additionally, an alternative configuration employing 30 labeled and 300 unlabeled cases for training was used for comparison.

**MRLiver:** The MRLiver dataset is collected in-house from UT Southwestern Medical Center. It contains 48 MRI scans, which include two types of sequences (T2 and T2_SPIR). Each scan includes 64 MRI slices, which cover the whole liver for segmentation. We randomly split the dataset into 30, 6, and 12 cases for training, validation, and testing. In the semi-supervised segmentation task, we used 1, 3, and 5 labeled scans and the corresponding 29, 27, and 25 unlabeled scans to form three semi-supervised training scenarios, respectively.

**3.5 Evaluation and implementation details**

All experiments were conducted on an NVIDIA 4090 GPU using a fixed batch size of 8, with a 50%/50% split between labeled and unlabeled data for each batch. The optimizer was set to SGD, with an initial learning rate of 0.01 for the Match-based network and 5e-5 for the SAM-based network. We performed a total of 60,000 iterations for SAMatch, consisting of 30,000 iterations for the warm-up stage, followed by 30,000 iterations of interactive training (Algorithm 1). All images were uniformly resized to 256 by 256 pixels before input into the network. All architectures, except for the SAM-based network, were based on U-Net for a fair comparison.

Our framework can integrate any SAM-based methods with existing Match-based methods. In this study, we adopted two representative Match-based methods: FixMatch and UniMatch. FixMatch is a representative work for Match-based methods and UniMatch is a state-of-the-art framework achieving top performance on three natural image segmentation datasets. For SAM-based methods, we adopted the original SAM network, and a SAM adaptation specifically tailored to medical images (MedSAM). We used points as prompts for SAM and boxes as prompts for MedSAM, according to their default settings. Point prompts were derived from the teacher model predictions of FixMatch or UniMatch, by selecting those with confidence values exceeding 0.95. For SAM, we randomly selected one positive point from the segmented objects and nine negative points from the background based on the pseudo-labels of the teacher model. In contrast, for MedSAM, we extracted a bounding box from the largest connected region of the pseudo-labels generated by the teacher model. This arrangement resulted in four variants from SAMatch: Fix-SAM, Fix-MedSAM, Uni-SAM, and Uni-MedSAM.

**3.6 Evaluation metrics**

For segmentation evaluation, we employ the DICE and the 95$^{th}$ Hausdorff Distance (HD95) as primary performance metrics. The DICE metric quantifies the overlap between the predicted and 'ground-truth' segmentation labels, defined as:

$$\text{DICE} = \frac{2|A \cap B|}{|A|+|B|}, \tag{5}$$

where A and B represent the predicted and actual segmentation regions, respectively. A DICE value (0-1) close to 1 indicates a high degree of overlap between the segmented regions. HD95 assesses the consistency of segmentation boundaries by measuring the $95^{th}$ percentile distance between them, thereby reducing the noise effects of outliers. HD95 is defined as:

$$\text{HD}_{95}(A, B) = \max\left(P_{a \in A} \min_{b \in B} d(a, b), P_{b \in B} \min_{a \in A} d(b, a)\right), \tag{6}$$

where $P$ means $95^{th}$ percentile, and $d(a, b)$ represents the Euclidean distance between boundary points a and b. Together, DICE and HD95 provide a comprehensive assessment of the model's segmentation accuracy.

# 4 Results

**4.1 Comparison with other methods**

In the ACDC experiments, various semi-supervised learning methods were evaluated for comparison with SAMatch. Specifically, DAN [42] and ADVENT [43] are based on adversarial learning. ICT [31], MT [29], UA-MT [30], and URPC [32] are Match-based methods without additional differential data augmentations. In contrast, U2PL, FixMatch, and UniMatch are Match-based approaches with differential data augmentations. Our four variants of SAMatch: Fix-SAM, Fix-MedSAM, Uni-SAM, and Uni-MedSAM demonstrate state-of-the-art performance, as shown in Table 3, when supervised with either one or three labeled cases. Notably, using just one labeled case yields significant improvements for SAMatch, highlighting the effectiveness of integrating SAM-based models into semi-supervised segmentation. Furthermore, when trained with three labeled cases, SAMatch remains highly competitive, with performance close to full supervision (UNet-fully) using the whole set of labeled training data (70). The performance of FixMatch and UniMatch both improved when integrated with SAM or MedSAM, indicating that SAMatch can be a general framework applicable to different variants of sub-components. Wilcoxon signed-rank tests showed that the differences between Uni-MedSAM and all other semi-supervised methods were statistically significant with p value less than 0.05 in terms of DICE and HD95.

**Table 3**. DICE and HD95 results for different semi-supervised segmentation methods on the ACDC dataset. All values were presented as mean ± s.d. The best results are highlighted in bold. The red color indicates the results from models using all labeled training data-70 cases. Statistical tests via Wilcoxon signed-rank tests found the differences between Uni-MedSAM and all other semi-supervised methods were statistically significant ($p<0.05$).

| Method | 1 labeled case (only end-diastolic) | | | | | | | |
|---|---|---|---|---|---|---|---|---|
| | RV | | MYO | | LV | | Mean DICE↑ | Mean HD95↓ |
| | DICE↑ | HD95↓ | DICE↑ | HD95↓ | DICE↑ | HD95↓ | | |
| UNet-Fully | 91.81±0.04 | 1.11±0.277 | 88.64±0.03 | 3.21±6.819 | 93.97±0.04 | 5.50±14.249 | 91.47±0.04 | 3.27±7.12 |
| DAN | 41.32±0.25 | 46.55±28.889 | 42.09±0.18 | 30.77±22.218 | 60.20±0.23 | 70.33±34.077 | 47.87±0.22 | 49.22±28.40 |
| ADVENT | 28.67±0.20 | 79.40±25.179 | 35.68±0.22 | 54.29±20.886 | 35.36±0.29 | 54.27±27.549 | 33.24±0.24 | 62.65±24.54 |
| ICT | 29.03±0.24 | 73.88±20.773 | 31.98±0.22 | 64.87±23.652 | 43.02±0.26 | 64.63±24.616 | 67.79±23.01 | 67.79±23.01 |
| MT | 31.23±0.26 | 46.72±26.984 | 34.93±0.25 | 37.55±32.678 | 41.59±0.32 | 39.59±27.743 | 41.29±29.14 | 41.29±29.14 |
| UA-MT | 22.90±0.24 | 39.92±27.795 | 33.73±0.25 | 41.71±25.369 | 40.19±0.28 | 53.17±29.818 | 44.93±27.66 | 44.93±27.66 |
| URPC | 24.15±0.26 | 49.09±33.002 | 27.23±0.24 | 42.95±25.591 | 35.95±0.29 | 72.10±25.480 | 29.11±0.26 | 54.71±28.02 |
| U2PL | 34.10±0.21 | 76.22±28.269 | 47.72±0.22 | 31.15±26.248 | 60.65±0.29 | 27.40±28.344 | 47.49±0.24 | 44.92±27.62 |
| FixMatch | 27.61±0.21 | 81.61±28.476 | 47.40±0.24 | 8.74±12.135 | 48.91±0.32 | 19.50±26.838 | 41.30±0.26 | 36.62±22.48 |
| Fix-SAM | 67.93±0.18 | 32.14±24.198 | 67.39±0.13 | 3.87±4.087 | 85.94±0.11 | 4.49±6.854 | 73.75±0.14 | 13.50±11.71 |
| Fix-MedSAM | 73.59±0.19 | 15.19±17.725 | 79.38±0.06 | 2.45±2.135 | 88.79±0.11 | 6.66±10.517 | 80.59±0.12 | 8.10±10.13 |
| UniMatch | 74.93±0.14 | 13.79±17.264 | 79.26±0.07 | **1.90±1.131** | 90.32±0.08 | 4.14±10.177 | 81.50±0.01 | 6.61±9.52 |
| Uni-SAM | 81.24±0.12 | 6.50±8.863 | 79.60±0.06 | 3.93±7.726 | **90.87±0.06** | **3.15±6.576** | 83.90±0.08 | 4.53±7.72 |
| Uni-MedSAM | **81.47±0.13** | **5.58±12.036** | **83.55±0.04** | 3.77±8.315 | 90.00±0.08 | 4.08±7.745 | **85.01±0.08** | **4.48±9.37** |
| Method | 3 labeled cases | | | | | | | |
| | RV | | MYO | | LV | | Mean DICE↑ | Mean HD95↓ |
| | DICE↑ | HD95↓ | DICE↑ | HD95↓ | DICE↑ | HD95↓ | | |
| UNet-Fully | 91.81±0.04 | 1.11±0.277 | 88.64±0.03 | 3.21±6.819 | 93.97±0.04 | 5.50±14.249 | 91.47±0.04 | 3.27±7.12 |
| DAN | 54.09±0.28 | 33.75±29.290 | 57.23±0.26 | 18.08±17.705 | 57.23±0.26 | 22.85±26.367 | 56.84±0.29 | 24.90±24.45 |
| ADVENT | 40.84±0.31 | 25.48±26.603 | 40.84±0.31 | 12.19±14.884 | 60.13±0.33 | 13.68±22.778 | 51.63±0.31 | 17.12±21.42 |
| ICT | 50.86±0.31 | 19.62±21.504 | 58.48±0.26 | 8.89±11.035 | 66.56±0.30 | 14.50±16.776 | 58.63±0.29 | 14.34±16.44 |
| MT | 44.30±0.30 | 40.25±29.632 | 54.73±0.26 | 11.83±14.531 | 63.67±0.30 | 20.91±26.348 | 54.23±0.29 | 24.33±23.50 |
| UA-MT | 26.30±22.18 | 55.76±34.401 | 50.55±0.26 | 10.16±11.845 | 55.96±0.34 | 12.97±20.303 | 49.47±0.20 | 26.30±22.18 |
| URPC | 42.07±0.32 | 25.61±27.837 | 46.73±0.28 | 10.36±13.219 | 46.73±0.28 | 12.35±15.960 | 48.46±0.32 | 16.11±19.01 |
| U2PL | 82.24±0.10 | 8.31±12.034 | 80.59±0.07 | 2.76±3.571 | 87.40±0.11 | 9.45±16.303 | 82.24±0.10 | 82.24±0.10 |
| FixMatch | 49.76±0.29 | 16.63±18.016 | 63.29±0.24 | 4.28±5.449 | 76.37±0.23 | 5.99±10.366 | 63.14±0.25 | 8.97±11.28 |
| Fix-SAM | 84.95±0.09 | 4.07±9.022 | 83.88±0.04 | 2.58±3.520 | 90.39±0.09 | 3.18±5.077 | 86.41±0.08 | 3.28±5.87 |
| Fix-MedSAM | 86.32±0.07 | 2.88±5.199 | 84.44±0.05 | 9.15±18.571 | 89.70±0.10 | 11.44±18.269 | 86.82±0.07 | 7.82±14.02 |
| UniMatch | 87.96±0.07 | 1.49±1.078 | 85.08±0.04 | 1.58±0.952 | 91.21±0.07 | 4.46±10.000 | 88.08±0.06 | 2.51±4.01 |
| Uni-SAM | 88.95±0.06 | 1.40±0.948 | 85.45±0.04 | **1.39±0.678** | 91.55±0.07 | **2.79±6.741** | 88.65±0.06 | **1.86±2.79** |
| Uni-MedSAM | **89.60±0.07** | **1.26±0.630** | **86.45±0.04** | 2.09±5.293 | **92.04±0.06** | 3.70±8.062 | **89.36±0.06** | 2.35±4.66 |

Figure 5 illustrates the class activation maps for segmented objects from the ACDC testing set, comparing MT, UA-MT, U2PL, FixMatch, and Fix-SAM. Notably, Fix-SAM focuses more on the 'ground-truth' objects than the other methods. Additionally, Fix-SAM shows smoother transitions near the boundaries of the segmented objects.

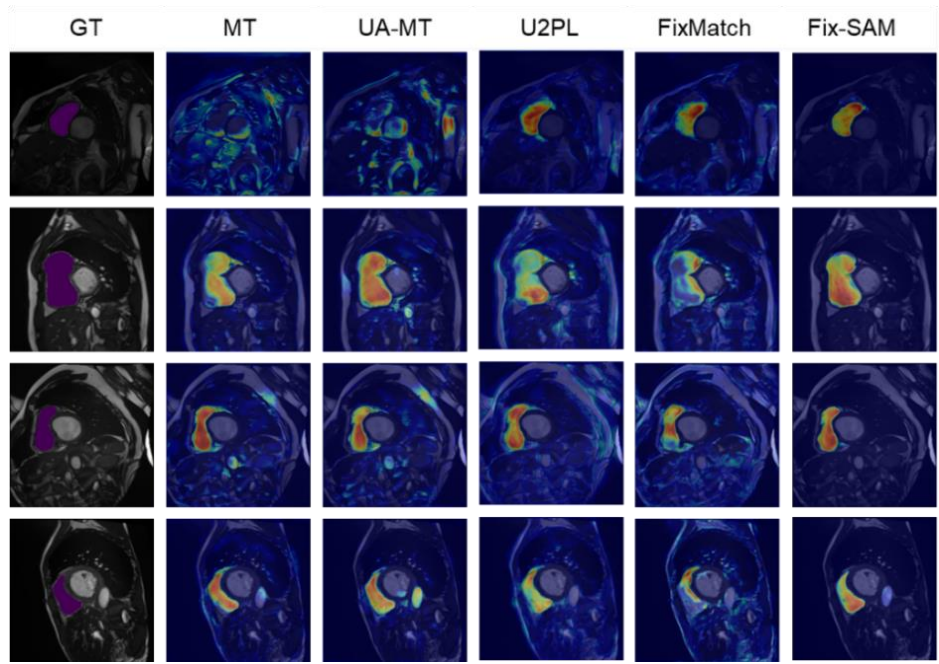

**Figure 5**. Class activation map results by using different semi-supervised segmentation methods.

Figure 6 presents segmentation examples on ACDC from FixMatch, UniMatch, and four variants of SAMatch. Compared to the two original Match-based methods, SAMatch generates segmentation results more consistent with the 'ground truth'. Notably, there is a significant improvement over FixMatch, highlighting our framework's ability to enhance Match-based methods when integrated with SAM-based methods.

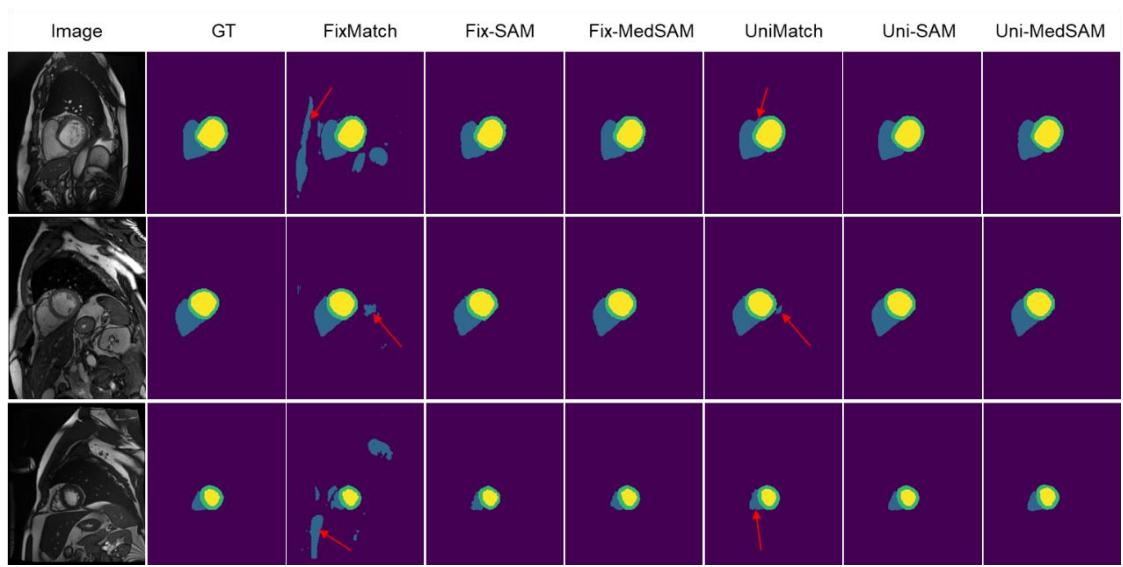

**Figure 6.** Visualization of segmentation results using different semi-supervised methods, trained with 3 labeled cases and 67 unlabeled cases.

In the BUSI dataset, we performed semi-supervised semantic segmentation training using 10 labeled images and 320 unlabeled images, as well as 30 labeled images and 300 unlabeled images. For SAMatch, we did not use MedSAM as a SAM-based variant, since MedSAM has been trained on this dataset as well and may bias the results. Table 4 compares various semi-supervised methods on the BUSI dataset, demonstrating that our method achieves state-of-the-art results. Notably, with just 30 labels, Uni-SAM closely matches the fully supervised method trained with 330 labels, showing only a 1.2% difference in DICE (in absolute terms, same in following texts). Furthermore, the minimal performance gap between training with 10 and 30 labels highlights our method's advantage in scenarios where labeled data is scarce.

**Table 4**. DICE scores for different semi-supervised segmentation methods on the BUSI Dataset. All values were presented as mean±s.d. The best results are highlighted in bold. The red color indicates the results from models using all labeled training data-330 cases. Wilcoxon signed-rank tests revealed statistically significant differences ($p < 0.05$) between Uni-SAM and all other semi-supervised methods, except for UniMatch trained with 10 labeled cases, which yielded a p-value of 0.22.

| Method | 10 labeled | | 30 labeled | |
| --- | --- | --- | --- | --- |
| | background | object | background | object |
| UNet-Fully | 96.30±0.046 | 61.69±0.348 | 96.30±0.046 | 61.69±0.348 |
| DAN | 94.09 ± 0.005 | 9.14 ± 0.040 | 94.48 ± 0.005 | 24.53 ± 0.112 |
| ADVENT | 94.20 ± 0.005 | 12.67 ± 0.067 | 94.21 ± 0.005 | 12.77 ± 0.080 |
| ICT | 94.20 ± 0.004 | 10.55 ± 0.060 | 94.20 ± 0.005 | 11.50 ± 0.076 |
| MT | 94.12 ± 0.004 | 5.51 ± 0.030 | 94.22 ± 0.005 | 13.13 ± 0.081 |
| UA-MT | 94.22 ± 0.005 | 14.54 ± 0.078 | 94.22 ± 0.005 | 13.06 ± 0.082 |
| URPC | 94.27±0.066 | 11.51±0.271 | 94.48±0.067 | 22.17±0.338 |
| U2PL | 94.43±0.067 | 23.85±0.334 | 94.64±0.067 | 26.54±0.338 |
| FixMatch | 94.34±0.067 | 20.38±0.299 | 94.80±0.066 | 34.95±0.354 |
| Fix-SAM | 94.38 ± 0.005 | 25.04 ± 0.112 | 94.84 ± 0.004 | 38.29 ± 0.131 |
| UniMatch | **96.03 ± 0.002** | 54.01 ± 0.127 | 95.62 ± 0.002 | 56.42 ± 0.103 |
| Uni-SAM | 95.99 ± 0.002 | **57.13 ± 0.119** | **96.17 ± 0.002** | **59.35 ± 0.116** |

The segmentation results on the BUSI dataset, utilizing 30 labeled training samples, demonstrate that our Uni-SAM method achieves a high degree of consistency with the 'ground truth', as illustrated in Figure 7.

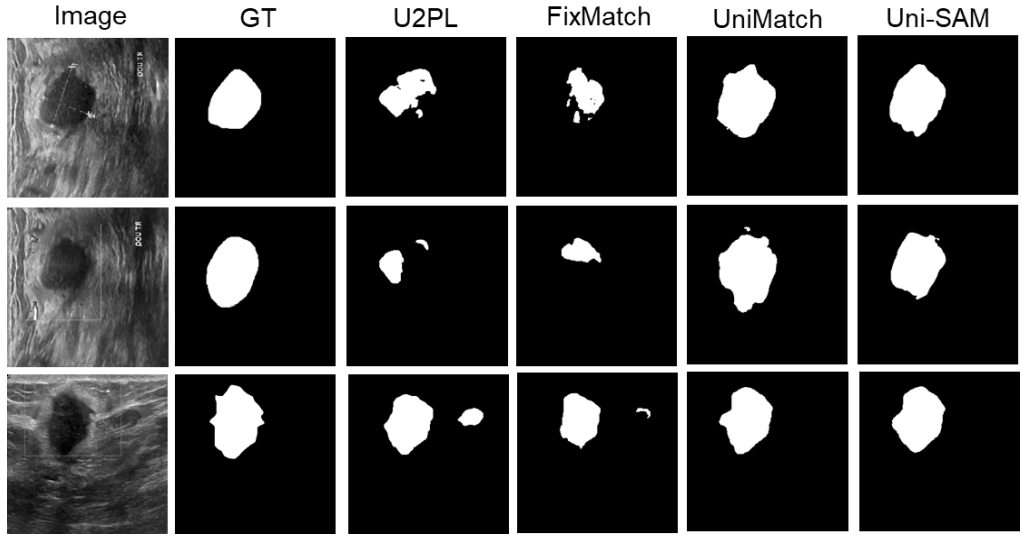

**Figure 7**. Segmentation results on BUSI with 30 labeled samples for training. The images of the first and second rows show benign lesions, and the third row shows a malignant lesion.

Results of the in-house MRLiver dataset are shown in Table 5. We observed consistent results with those of the ACDC and BUSI datasets. The performance of FixMatch and UniMatch improved significantly when integrated with the SAM or MedSAM models. For FixMatch, the average DICE scores increased by approximately 5% and 12% with SAM and MedSAM, respectively. For UniMatch, the corresponding boosts are 9% and 8%.

**Table 5**. DICE score and HD95 results for different semi-supervised segmentation methods on the MRLiver Dataset. All values were presented as mean±st.d. The best results are highlighted in bold. The red color indicates the results from models using all labeled training data-30 cases. Wilcoxon signed-rank tests showed statistically significant differences ($p < 0.05$) between Uni-MedSAM and all other semi-supervised methods, except for Uni-SAM trained with 5 labeled cases, which had a p-value of 0.39.

| Method | 1 Case | | 3 Cases | | 5 Cases | | Mean DICE↑ | Mean HD95↓ |
|---|---|---|---|---|---|---|---|---|
| | DICE↑ | HD95↓ | DICE↑ | HD95↓ | DICE↑ | HD95↓ | | |
| UNet-Fully | 79.45±0.120 | 10.37±8.717 | 79.45±0.120 | 10.37±8.717 | 79.45±0.120 | 10.37±8.717 | 79.45±0.120 | 10.37±8.717 |
| DAN | 49.93±0.258 | 37.57±24.273 | 66.66±0.192 | 36.78±21.868 | 70.41±0.139 | 29.47±24.021 | 62.33±0.196 | 34.61±23.387 |
| ADVENT | 46.50±0.311 | 37.88±25.125 | 68.72±0.123 | 37.14±23.564 | 67.30±0.155 | 33.60±25.085 | 60.84±0.196 | 36.21±24.591 |
| ICT | 47.98±0.289 | 39.65±27.711 | 70.07±0.122 | 33.04±23.337 | 69.89±0.143 | 31.28±25.193 | 62.65±0.185 | 34.66±25.414 |
| MT | 50.03±0.217 | 49.97±27.214 | 66.07±0.202 | 33.56±21.953 | 64.17±0.204 | 32.39±25.207 | 60.09±0.208 | 38.64±24.791 |
| UA-MT | 52.33±0.241 | 42.13±23.881 | 68.27±0.155 | 35.76±23.390 | 66.53±0.154 | 29.51±24.934 | 62.38±0.183 | 35.8±24.068 |
| URPC | 45.28±0.317 | 49.89±26.707 | 36.69±0.169 | 65.22±23.271 | 47.75±0.200 | 82.84±11.695 | 43.24±0.229 | 65.98±20.558 |
| U2PL | 41.43±0.318 | 37.89±36.670 | 64.56±0.184 | 27.14±20.277 | 69.95±0.107 | 29.12±24.912 | 58.65±0.203 | 31.38±27.286 |
| FixMatch | 48.37±0.298 | 44.55±21.578 | 60.38±0.235 | 27.09±19.793 | 63.82±0.216 | 22.58±19.512 | 57.52±0.25 | 31.41±20.294 |
| Fix-SAM | 52.17±0.204 | 52.21±21.421 | 67.07±0.105 | 31.24±23.384 | 69.12±0.140 | 20.51±17.401 | 62.79±0.15 | 34.65±20.735 |
| Fix-MedSAM | 58.34±0.122 | 37.01±20.866 | 73.44±0.128 | 34.12±20.841 | 70.55±0.113 | 27.21±22.945 | 67.44±0.121 | 32.78±21.551 |
| UniMatch | 46.94±0.339 | 44.26±24.618 | 74.96±0.144 | 21.78±21.624 | 75.86±0.116 | 22.28±22.082 | 65.92±0.2 | 29.44±22.275 |
| Uni-SAM | **68.11±0.094** | **25.24±14.286** | 77.13±0.097 | 31.29±20.793 | **79.86±0.076** | **15.35±13.383** | **75.03±0.089** | 23.96±16.154 |
| Uni-MedSAM | 65.46±0.068 | 28.74±12.044 | **80.04±0.111** | 21.04±21.983 | 78.75±0.095 | 20.16±20.636 | 74.75±0.091 | **23.31±18.221** |

In Figure 8, we present the liver segmentation results derived from models trained with 3 labeled and 27 unlabeled cases. Compared to FixMatch and UniMatch, the segmentation results from the four variants of SAMatch demonstrate various degrees of improvement. However, Fix-SAM and Uni-SAM are prone to over-segmentation. This issue may be attributed to the point prompts used for SAM, while MedSAM employs bounding boxes as prompts, providing a stronger constraint for segmentation.

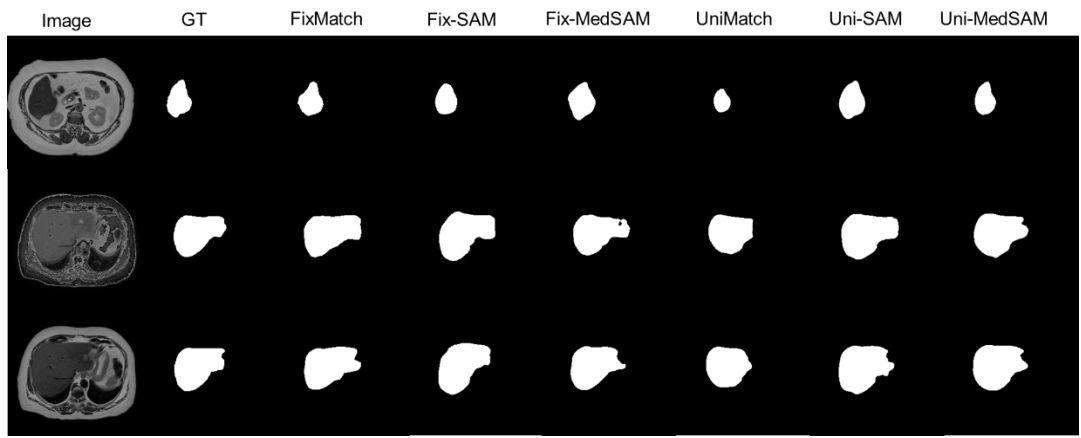

**Figure 8**. Segmentation results on the MRliver testing dataset. These results are based on models trained with 3 labeled cases and 27 unlabeled cases.

## 5. Discussion

In this study, we propose a unified semi-supervised framework, SAMatch, for medical image segmentation. SAMatch integrates the strengths of the foundational segmentation model SAM with those of a Match-based framework. SAMatch allows the integration of different variants of SAM-based methods (SAM and MedSAM for this study) with different Match-based architectures (FixMatch and UniMatch for this study), which can be tailored to the specific needs of segmentation tasks. As shown, SAMatch consistently improves the accuracy of the original Match-based frameworks, and significantly outperforms other comparison methods (Table 3-5, Figure 5-8), offering a general solution. As expected, the UniMatch framework generated better results than the FixMatch framework, and the corresponding SAMatch results based on UniMatch also outperformed those based on FixMatch. MedSAM, fine-tuned on a large dataset of medical images, also showed some advantages over the vanilla SAM framework when used in SAMatch. The flexibility of SAMatch allows future, more advanced Match-based methods, and SAM-based methods to be integrated to maximize the segmentation accuracy.

The SAMatch framework can be further improved in future studies, mostly in three aspects:

(1) Prompts for SAM-based models

Experimental results demonstrate that the Match-based model can automatically generate prompts for SAM-based methods, which enables further pseudo-label fine-tuning. However, we observed instances of misalignment in the generated prompts. For example, point prompts may correspond to background regions while being misclassified as objects in the prediction masks of the Match-based models. When comparing the prediction results from SAM using prompts extracted from 'ground-truth' labels to those based on the teacher model prompts, we observe a clear performance gap that needs to be addressed. To improve prompt accuracy, incorporating constraint conditions, such as object shape or boundary regularizations for Match-based models, may further enhance the accuracy of segmentation tasks.

(2) Coupled training and knowledge-distillation

Our SAMatch framework could alleviate the coupling effect between the teacher and student models with the introduction of the SAM-based method. Specifically, the SAM-based method functions as a teacher model assistant that generates pseudo-labels for training the student model. Experimental results demonstrate that SAM-based methods exhibit greater robustness and accuracy in pseudo-label generation when provided with appropriate prompts. In future studies, we can extract representative features from SAM, which may facilitate the transfer of knowledge to the Match-based model during the training phase. For instance, features from SAM can be concatenated with those from the Match-based models interactively, potentially enhancing the final segmentation performance of the Match-based models. Additionally, the development of efficient, compact Match-based models through the integration of knowledge distillation strategies, such as pruning [44] or quantization [45], can further enhance SAMatch.

(3) 3D segmentation

In Table 3, we observed relatively low accuracy in lesion segmentation, primarily due to the low contrast between lesions and surrounding tissues, as well as significant shape variations across different lesions. These factors present a challenge for semi-supervised models to achieve accurate lesion segmentation based on limited labeled data. As our current model is implemented in 2D, extending it to 3D could help leverage spatial shape continuities in another dimension to improve segmentation accuracy by more effectively distinguishing complex lesions. Additionally, recent advancements in 3D and video segmentation, such as SAM 2 [46] and MedSAM-2 [47], demonstrate the potential to enhance performance in segmentation tasks, including lesion segmentation, and could offer valuable insights for improving semi-supervised learning in medical imaging.

## 6. Conclusion

In this study, we proposed a semi-supervised framework SAMatch for medical image segmentation. It combines a Match-based framework with a foundational segmentation model to harness the generalizability advantage of the latter to refine the intermediate pseudo-labels generated by the Match-based framework, which helps to enforce consistency regularization-driven Match-based framework training. Experimental results demonstrated that SAMatch promotes effective semi-supervised learning through limited labeled data, which can especially benefit clinical scenarios with limited human annotations.

## Acknowledgements

We would like to thank Yogesh Rathi, Lipeng Ning, and Jun Lyu for their insightful and helpful discussions.